\documentclass{article}

\usepackage{microtype}
\usepackage{graphicx}
\usepackage{subcaption}
\usepackage{booktabs}
\usepackage{tabularx}
\usepackage{hyperref}

\usepackage[accepted]{icml2026}
\icmlshowauthorstrue

\makeatletter
\renewcommand{\Notice@String}{\textit{Accepted to the ICML 2026 Workshop on
Efficient Multimodal Question Answering (EMM-QA)}, Seoul, South Korea. Copyright 2026 by the author(s).}
\makeatother

\usepackage{amsmath}
\usepackage{amssymb}
\usepackage{mathtools}
\usepackage{amsthm}
\usepackage{xcolor}
\usepackage{url}
\usepackage[capitalize,noabbrev]{cleveref}
\usepackage{placeins}

\graphicspath{{../ICML_2026___DCVLR___Igor/dcvlr_paper_icml/}}

\theoremstyle{plain}

\theoremstyle{definition}

\theoremstyle{remark}

\icmltitlerunning{Data-Efficient Curation for Multimodal Reasoning under Fixed Training Protocols}

\begin{document}

\twocolumn[
\icmltitle{Data-Efficient Curation for Multimodal Reasoning under Fixed Training Protocols}

\icmlsetsymbol{equal}{*}

\begin{icmlauthorlist}
\icmlauthor{Yosub Shin}{equal,uhm}
\icmlauthor{Michael Buriek}{equal,uhm,pwc}
\icmlauthor{Boris Sobolev}{equal,uhm,cisco}
\icmlauthor{Pavel Bushuyeu}{equal,uhm}
\icmlauthor{Vikas Kumar}{uhm}
\icmlauthor{Haoyang Xu}{uhm}
\icmlauthor{Samuel Watson}{uhm}
\icmlauthor{Igor Molybog}{uhm}
\end{icmlauthorlist}

\icmlaffiliation{uhm}{University of Hawai'i at M\=anoa, Honolulu, HI, USA}
\icmlaffiliation{pwc}{PwC, USA}
\icmlaffiliation{cisco}{Cisco, USA}

\icmlcorrespondingauthor{Yosub Shin}{yosubs@hawaii.edu}

\icmlkeywords{Data Curation, Vision-Language Reasoning, Data Efficiency, Multimodal QA, Machine Learning}

\vskip 0.3in
]

\printAffiliationsAndNotice{\icmlEqualContribution}

\begin{abstract}
We study data curation for multimodal reasoning in a fixed-protocol fine-tuning regime, where the base model, optimizer, training schedule, and evaluation pipeline are held constant and the main degree of freedom is the training data. Using the NeurIPS 2025 Data Curation for Vision--Language Reasoning (DCVLR) challenge~\citep{dcvlr2025website} as a controlled testbed, we analyze how source-dataset alignment, model-relative difficulty, dataset size, diversity heuristics, and rewritten synthetic mixtures affect downstream reasoning accuracy. Among the tested interventions, difficulty filtering on an aligned source corpus provides the strongest gains at matched scale. The effect is not explained only by LiveXivTQA weighting: a per-benchmark decomposition shows that much of the improvement over random sampling comes from OlympiadBench, the largest non-LiveXivTQA benchmark. Qwen-derived difficulty scores also transfer to some additional model families, though the benefit is architecture-dependent. In contrast, increasing dataset size beyond roughly 1k aligned examples mainly reduces run-to-run variance under the fixed recipe, while the diversity and rewritten CoSyn mixtures we tested do not improve over the difficulty-filtered baseline. These results provide a scoped empirical recipe for data-constrained multimodal reasoning fine-tuning, rather than a universal claim about data selection across all training regimes.
\end{abstract}

\section{Introduction}
\label{sec:intro}

Data curation is increasingly central to multimodal question answering and reasoning, but its effects are hard to isolate. Many reported gains in vision-language reasoning combine changes in model architecture, training recipe, supervision style, and data composition, making it unclear which part of the pipeline explains downstream improvement. This paper studies a narrower question: when the model and fine-tuning protocol are fixed, which data curation choices matter most?

We examine this question through the NeurIPS 2025 DCVLR challenge. DCVLR fixes the base model and training procedure and varies only the curated training set, making it a restrictive but useful testbed for studying data-efficient multimodal reasoning under realistic resource constraints. Our goal is not to propose a new model architecture or a universal data-selection algorithm. Instead, we provide a controlled empirical characterization of which curation signals are effective in this fixed-protocol regime.

Our 1k-example submission, derived primarily from the Walton Multimodal Cold Start corpus~\citep{oumi2025waltondataset}, improved the aggregate DCVLR score from 38.4 to 46.0. We use this outcome as a starting point, then run post-competition ablations to identify which choices explain the gain. The main pattern is simple: choosing an aligned source corpus is a prerequisite, and within that source, filtering for examples that are challenging but learnable is the strongest signal among the interventions we tested.

We make four contributions. First, we use DCVLR as a controlled testbed for data-efficient multimodal reasoning, where data selection can be studied without confounding changes in model or optimizer. Second, we show that model-relative difficulty filtering on an aligned source dataset outperforms random, easy-only, and stricter hard-only selection at matched scale. Third, we provide per-benchmark, weighting-decomposition, and cross-model analyses showing that the gain is not solely a LiveXivTQA weighting artifact or a Qwen-only artifact. Fourth, we report negative results for representative diversity and rewritten synthetic-data mixtures, clarifying that these heuristics did not add benefit in the tested fixed-recipe regime.

\section{Setting and Curation Method}
\label{sec:method}

\paragraph{DCVLR as a fixed-protocol benchmark.}
The DCVLR evaluation suite contains ten benchmarks spanning text-heavy academic QA, mathematics, physics, and general visual reasoning~\citep{dcvlr2025resultblog}. Some benchmarks, including LiveXivTQA~\citep{shabtay2025livexivmultimodallive} and VMCBench-DEV~\citep{zhang2025vmcbench}, were known during development; others, including Omni3DBench~\citep{marsili2025omni3dbench} and the Yale Physics subsets~\citep{feng2025physics}, were held out until final evaluation. The organizers fixed the model, optimizer, learning-rate schedule, training duration, and evaluation pipeline. This makes DCVLR less informative about unconstrained training recipes, but unusually useful for isolating data effects.

\begin{figure}[t]
\centering
\includegraphics[width=0.95\linewidth]{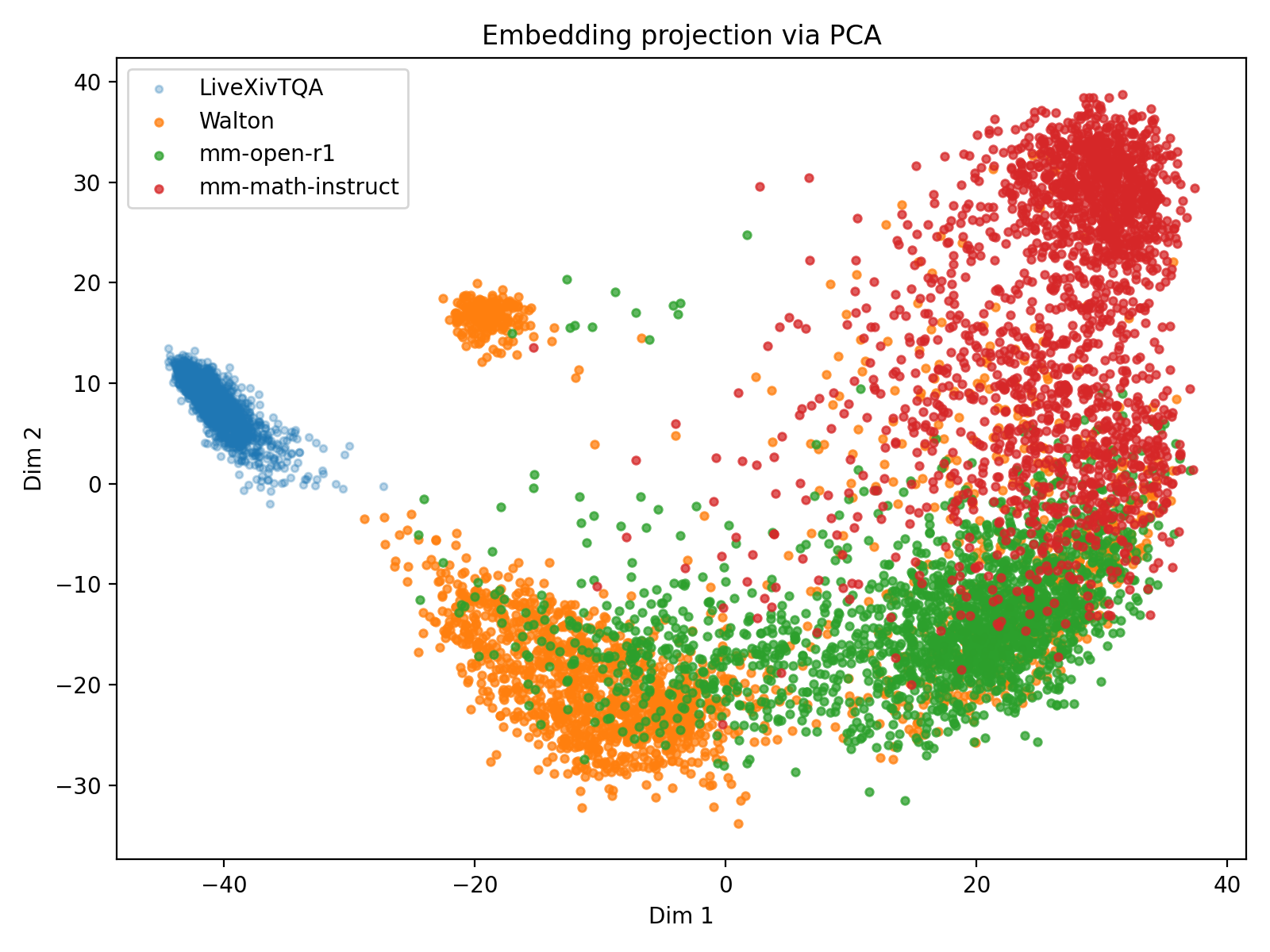}
\caption{PCA projection of Qwen2.5-VL-7B-Instruct embeddings for LiveXivTQA and three DCVLR baseline datasets. Walton lies closest to the LiveXivTQA cluster, motivating it as the aligned starting corpus for the controlled ablations.}
\label{fig:pca_embed}
\end{figure}

\paragraph{Starting from an aligned source.}
We begin from Walton Multimodal Cold Start~\citep{oumi2025waltondataset}, one of the organizer-provided datasets~\citep{oumi2025trainingfrontiervlms}. Walton was chosen because it is aligned with the dominant academic-QA portion of the benchmark and with the Qwen2.5-VL-7B-Instruct base model~\citep{bai2025qwen25vltechnicalreport}. Figure~\ref{fig:pca_embed} visualizes this alignment: in the Qwen2.5-VL representation space, Walton lies closer to LiveXivTQA than MM-Open-R1~\citep{oumi2025mmopenr1} or MM-MathInstruct~\citep{oumi2025mmmathinstruct}. The base model is also more accurate on LiveXivTQA questions whose nearest neighbors are Walton examples (Appendix~\ref{app:alignment}). We do not interpret this as evidence that Walton is universally preferable. Rather, it is the most aligned starting point among the tested organizer-provided sources for this fixed benchmark and base model.

\paragraph{Model-relative difficulty.}
For each Walton example, we query Qwen2.5-VL-7B-Instruct with stochastic decoding for 16 rollouts using temperature 0.7 and top-$p$ 0.9. Let $k$ denote the number of correct final answers out of 16. High-$k$ examples are already solved reliably by the base model and are treated as easy; low-$k$ examples expose failures or instability and are treated as difficult. Each rollout is scored correct or incorrect with the same staged matcher used in our DCVLR evaluation pipeline (Appendix~\ref{app:impl}): exact case- and whitespace-normalized matching, then structured extraction (\verb|\boxed{}|, tags, and SymPy equivalence), then an LLM judge for unresolved cases; $k$ is the number of the 16 rollouts judged correct.
This is a model-relative definition of difficulty, not an intrinsic measure of human problem difficulty; characterizing difficulty by a base model's inconsistency under stochastic decoding has precedent in reasoning fine-tuning~\citep{gekhman2024finetuning}.

We use 16 rollouts for fine-grained analysis, but the main selection effect is coarse. Preliminary checks with 8 rollouts produced the same qualitative filtering behavior, and our primary selected region is a broad thresholded pool rather than a fine ranking. We therefore do not claim that 16 rollouts are necessary; lower rollout counts may be sufficient for coarse difficult-example selection.

\paragraph{Other curation interventions.}
We also test representative diversity and augmentation strategies: embedding-cluster balancing, category-level balancing, category exclusions, and mixtures with CoSyn-400K~\citep{yang2025scaling}. CoSyn examples are rewritten with GPT-4o into longer reasoning traces while preserving the original final answer, then mixed with Walton at controlled ratios. These interventions are evaluated as matched-scale ablations rather than assumed to compose additively with difficulty filtering. Figure~\ref{fig:pipeline} summarizes the pipeline, and implementation details, prompts, and seed counts appear in Appendix~\ref{app:impl}.

\begin{figure*}[t]
\centering
\includegraphics[width=0.94\textwidth]{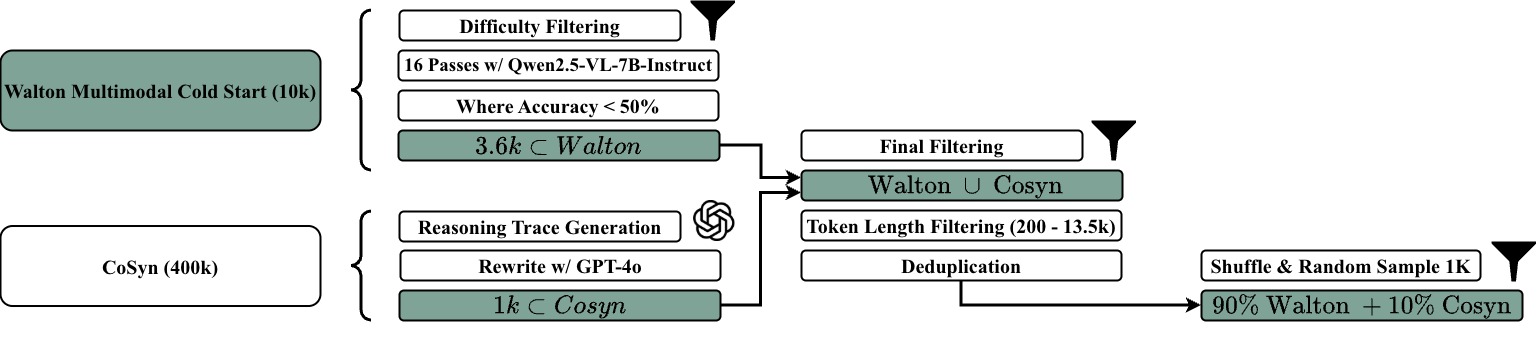}
\caption{Overview of the dataset curation pipeline, from candidate data sources through difficulty scoring, optional diversity or synthetic-data interventions, and final fixed-size sampling.}
\label{fig:pipeline}
\end{figure*}

\section{Experimental Setup}
\label{sec:exp}

The main experiments fine-tune Qwen2.5-VL-7B-Instruct using the official DCVLR training recipe. For robustness checks, we train Molmo-7B, Phi-4, and Gemma3-4B on matched random and difficulty-filtered subsets selected using the Qwen-derived scores. Models are evaluated on the full DCVLR suite, and we report both aggregate accuracy and per-benchmark accuracy where relevant.

We use repeated runs where compute permitted. The random 1k Walton baseline uses five repetitions; difficulty-threshold ablations use three repetitions per threshold; dataset-size ablations use three to five repetitions depending on size, except the 10k endpoint. Cross-model, CoSyn-mixture, and most diversity ablations should be interpreted as robustness or exploratory checks unless otherwise noted. Appendix~\ref{app:impl} lists the repetition counts and compute budget.

\section{Results}
\label{sec:results}

\subsection{Competition Outcome}
\label{sec:outcome}

Table~\ref{tab:main} reports per-benchmark accuracy for the base model and our curated 1k submission. The aggregate DCVLR score improves from 38.4 to 46.0. The large gain on LiveXivTQA ($56.8 \to 74.5$) is consistent with Walton's alignment to that benchmark, while smaller or negative deltas on held-out benchmarks reflect the specialization--generalization trade-off analyzed below. We use this outcome as the starting point for the ablations that follow.

\begin{table}[t]
\centering
\caption{Per-benchmark accuracy (\%) for the base model and our curated 1k submission. This establishes the competition outcome; the ablations in \cref{sec:difficulty,sec:size,sec:diversity_results} explain it.}
\label{tab:main}
\scriptsize
\setlength{\tabcolsep}{4pt}
\begin{tabular}{lrrr}
\toprule
Benchmark & \# Samples & Base & Ours 1k \\
\midrule
VMCBench DEV    & 1000  & 79.8 & 78.2 \\
LiveXivTQA      & 7913  & 56.8 & 74.5 \\
OlympiadBench   & 5929  & 13.4 & 11.7 \\
Omni3DBench     & 501   & 34.8 & 32.6 \\
Atomic          & 200   & 8.5  & 11.5 \\
Electro         & 242   & 8.3  & 7.4  \\
Mechanics       & 221   & 9.0  & 8.1  \\
Optics          & 158   & 7.0  & 8.9  \\
Quantum         & 236   & 5.9  & 6.8  \\
Statistics      & 240   & 17.1 & 14.2 \\
\midrule
Overall (weighted) & 16640 & 38.4 & 46.0 \\
\bottomrule
\end{tabular}
\end{table}

\subsection{Difficulty Filtering Is the Strongest Tested Signal}
\label{sec:difficulty}

Table~\ref{tab:walton_difficulty_ablation} compares thresholded Walton subsets at fixed size. The rows are thresholded pools, not disjoint bins: for example, the moderate threshold $k \le 8$ includes examples that also satisfy stricter thresholds. The moderate threshold outperforms random sampling, easy examples, and the stricter hard-only threshold. This suggests that the useful signal is not simply model failure: examples that are too easy provide little learning signal, while examples that are nearly always failed can be unstable or poorly matched to the fixed recipe.

\begin{table}[h]
\centering
\caption{Difficulty-threshold ablation on Walton. Each subset contains 1k examples. Accuracy is aggregate DCVLR accuracy as a fraction.}
\label{tab:walton_difficulty_ablation}
\setlength{\tabcolsep}{4pt}
\begin{tabular}{lcc}
\toprule
Subset & Threshold & Mean $\pm$ Std \\
\midrule
Hard threshold       & $k \le 3$  & $0.472 \pm 0.023$ \\
Moderate threshold   & $k \le 8$  & $\mathbf{0.491 \pm 0.002}$ \\
Easy threshold       & $k \ge 15$ & $0.440 \pm 0.024$ \\
Random Walton        & --         & $0.475 \pm 0.010$ \\
\bottomrule
\end{tabular}
\end{table}

\paragraph{Not only LiveXivTQA weighting.}
Table~\ref{tab:benchmark_breakdown} compares random 1k Walton subsets with the moderate-threshold subset. Difficulty filtering improves the aligned LiveXivTQA benchmark, but the larger absolute gain over random sampling occurs on OlympiadBench, the largest non-LiveXivTQA benchmark. A simple weighting decomposition of the overall improvement over random sampling ($+0.016$) assigns approximately $+0.004$ to LiveXivTQA and $+0.011$ to OlympiadBench. Thus, most of the aggregate difference between random and difficulty-filtered 1k subsets is not produced by LiveXivTQA alone.

\begin{table}[h]
\centering
\caption{Per-benchmark comparison of random 1k Walton vs. the difficulty-filtered subset. Difficulty filtering improves the largest non-LiveXivTQA benchmark, reducing the concern that the aggregate gain is only a weighting artifact.}
\label{tab:benchmark_breakdown}
\scriptsize
\setlength{\tabcolsep}{3pt}
\begin{tabular}{lccc}
\toprule
Benchmark & Base & Random 1k & $k \le 8$ \\
\midrule
LiveXivTQA        & 0.568 & $0.779 \pm 0.011$ & $\mathbf{0.788 \pm 0.002}$ \\
OlympiadBench     & 0.134 & $0.112 \pm 0.009$ & $\mathbf{0.144 \pm 0.002}$ \\
Physics           & 0.095 & $0.091 \pm 0.016$ & $0.098 \pm 0.007$ \\
VMCBench-DEV      & 0.798 & $0.787 \pm 0.006$ & $0.790 \pm 0.003$ \\
Omni3DBench       & 0.348 & $0.341 \pm 0.005$ & $0.341 \pm 0.003$ \\
\midrule
Overall           & 0.384 & $0.475 \pm 0.010$ & $\mathbf{0.491 \pm 0.002}$ \\
\bottomrule
\end{tabular}
\end{table}

\paragraph{Cross-model transfer is partial.}
To test whether the signal is purely Qwen-specific, we train additional model families on subsets selected using the same Qwen-derived scores (Table~\ref{tab:cross_model}). The difficulty-filtered subset improves Molmo-7B and Phi-4 relative to random subsets. Gemma3-4B does not improve ($0.307$ vs.\ $0.316$ for random). As a further check that the signal is not tied to the scoring model, training Phi-4 on data scored by Phi-4 itself reaches 0.331, close to the 0.327 obtained with Qwen-derived scores. We therefore interpret Qwen-derived difficulty as a transferable but model-dependent signal, not as an architecture-invariant definition of task difficulty.

\begin{table}[t]
\centering
\caption{Cross-model transfer. Difficulty scores are computed with Qwen2.5-VL-7B-Instruct, while the training targets are different model families. These are robustness checks, not exhaustive multi-seed studies.}
\label{tab:cross_model}
\setlength{\tabcolsep}{5pt}
\begin{tabular}{lcc}
\toprule
Target model & Random 1k & $k \le 8$ subset \\
\midrule
Molmo-7B    & 0.182 & \textbf{0.233} \\
Phi-4       & 0.314 & \textbf{0.327} \\
Gemma3-4B   & \textbf{0.316} & 0.307 \\
\bottomrule
\end{tabular}
\end{table}

\subsection{Dataset Size Mostly Affects Stability}
\label{sec:size}

\begin{figure}[t]
\centering
\includegraphics[width=0.95\linewidth]{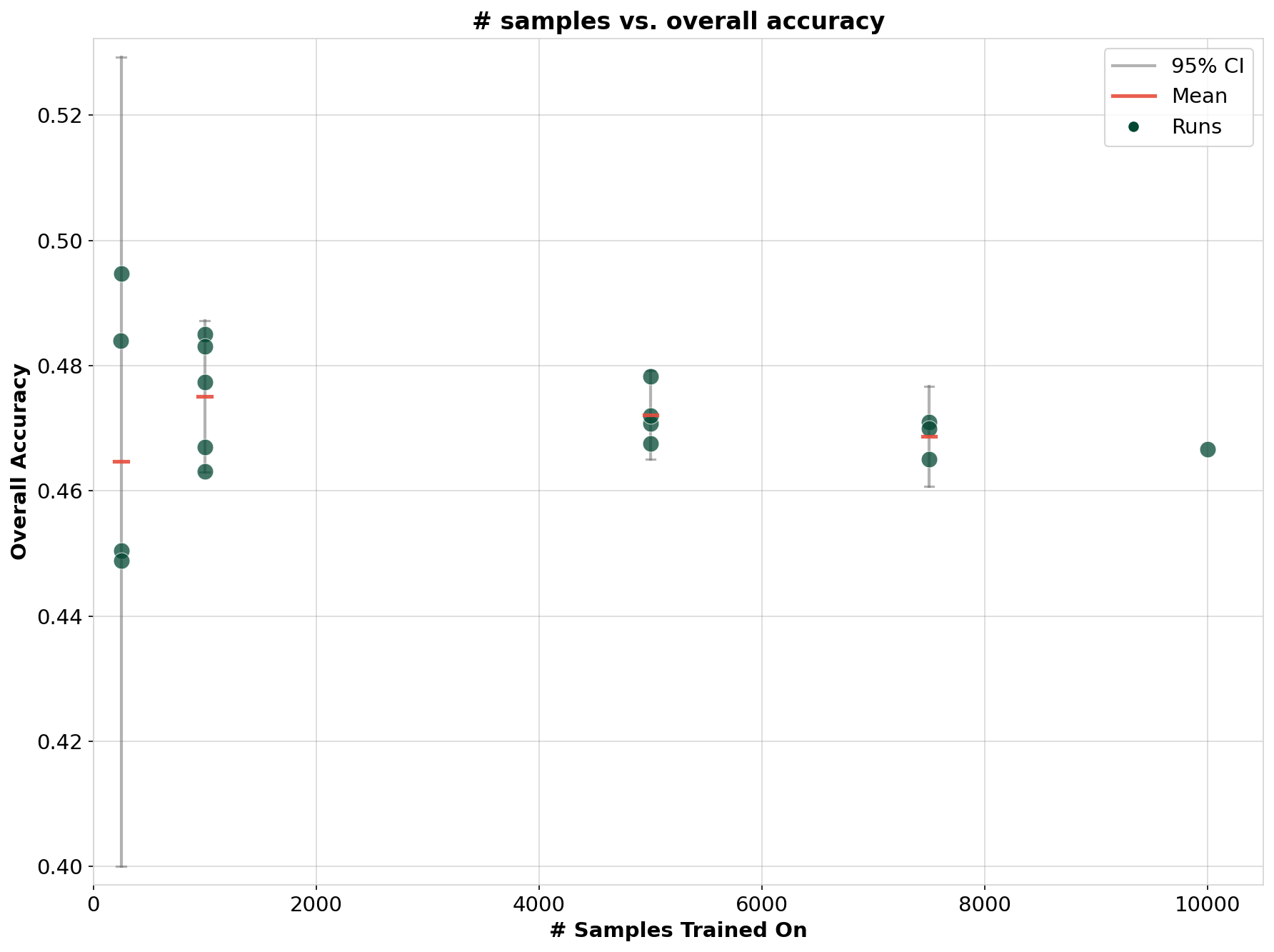}
\caption{Overall accuracy vs. dataset size for randomly sampled Walton subsets. Mean accuracy plateaus around 1k examples under the fixed training recipe, while variance decreases with scale.}
\label{fig:dataset_size_trends}
\end{figure}

Dataset size behaves differently from difficulty. Moving from very small subsets to roughly 1k examples yields most of the mean-performance gain. Beyond this point, increasing the number of randomly sampled Walton examples does not reliably improve aggregate accuracy under the fixed recipe: overall accuracy is $0.475 \pm 0.008$ at 1k, $0.472 \pm 0.004$ at 5k, $0.469 \pm 0.003$ at 7.5k, and 0.467 at 10k. Larger datasets reduce run-to-run variance, but the per-benchmark trends show no consistent evidence that scale alone expands transfer in this setting (Figure~\ref{fig:dataset_size_trends}).

\begin{figure*}[t]
\centering
\begin{subfigure}[t]{0.49\linewidth}
  \centering
  \includegraphics[width=\linewidth]{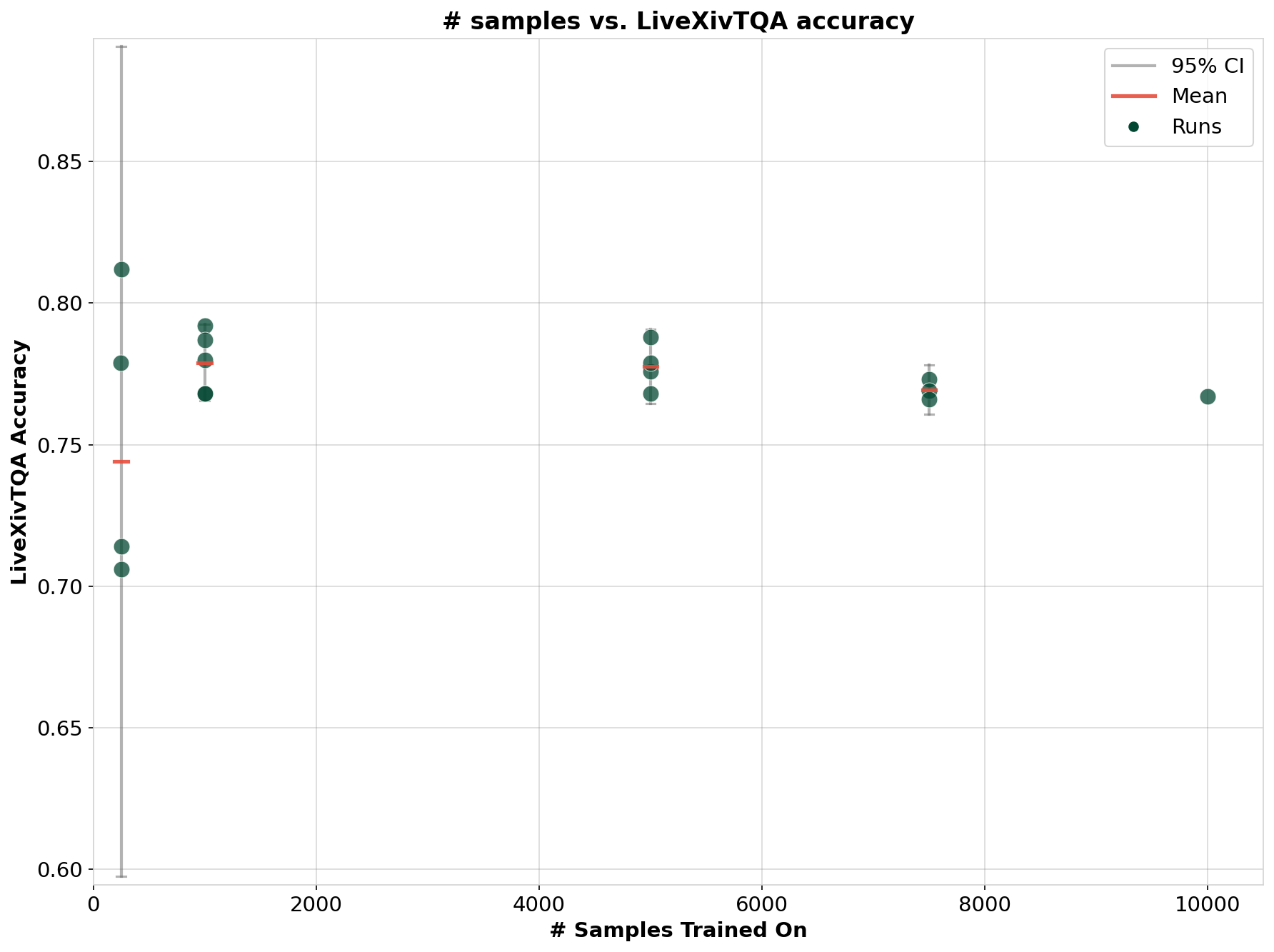}
  \caption{LiveXivTQA}
\end{subfigure}
\hfill
\begin{subfigure}[t]{0.49\linewidth}
  \centering
  \includegraphics[width=\linewidth]{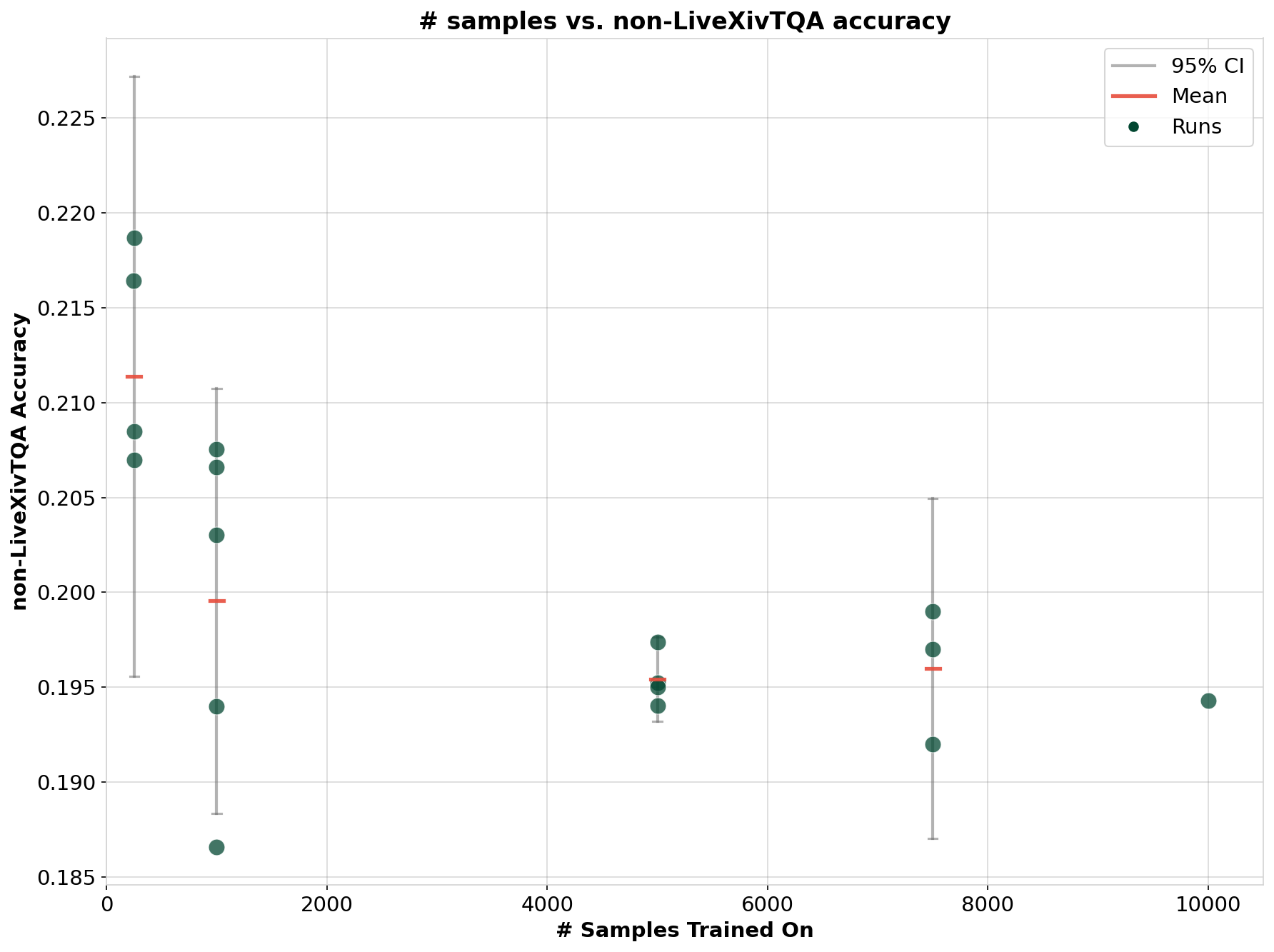}
  \caption{Non-LiveXivTQA}
\end{subfigure}
\caption{Accuracy vs. dataset size for the aligned LiveXivTQA benchmark and aggregate non-LiveXivTQA benchmarks.}
\label{fig:dataset_size_trends_appendix}
\end{figure*}

The per-benchmark pattern is consistent with fixed-recipe specialization, but should be interpreted cautiously. LiveXivTQA remains high after the initial gain and then plateaus, while OlympiadBench and Physics show mild degradation as more aligned Walton data is added (Figure~\ref{fig:dataset_size_trends_appendix}). Because confidence intervals overlap in several cases, we treat this as suggestive evidence of specialization rather than a definitive monotonic trade-off. The full size-by-benchmark breakdown is given in Table~\ref{tab:size_breakdown}.

\begin{table*}[t]
\centering
\caption{Per-benchmark dataset-size ablation on randomly sampled Walton subsets. Values are accuracy fractions. The pattern suggests fixed-recipe saturation: mean aggregate accuracy plateaus, while larger datasets reduce variance.}
\label{tab:size_breakdown}
\scriptsize
\setlength{\tabcolsep}{3pt}
\begin{tabular}{lcccccc}
\toprule
Benchmark & Base & 250 & 1k & 5k & 7.5k & 10k \\
\midrule
LiveXivTQA & 0.568 & $0.753 \pm 0.044$ & $0.779 \pm 0.010$ & $0.777 \pm 0.009$ & $0.769 \pm 0.004$ & 0.767 \\
OlympiadBench & 0.134 & $0.128 \pm 0.006$ & $0.112 \pm 0.009$ & $0.112 \pm 0.002$ & $0.112 \pm 0.005$ & 0.106 \\
Physics & 0.095 & $0.103 \pm 0.007$ & $0.091 \pm 0.016$ & $0.082 \pm 0.004$ & $0.080 \pm 0.007$ & 0.086 \\
VMCBench-DEV & 0.798 & $0.789 \pm 0.003$ & $0.787 \pm 0.006$ & $0.766 \pm 0.006$ & $0.770 \pm 0.009$ & 0.784 \\
Omni3DBench & 0.348 & $0.351 \pm 0.010$ & $0.341 \pm 0.005$ & $0.342 \pm 0.010$ & $0.340 \pm 0.003$ & 0.342 \\
\midrule
Overall & 0.384 & $0.470 \pm 0.020$ & $0.475 \pm 0.008$ & $0.472 \pm 0.004$ & $0.469 \pm 0.003$ & 0.467 \\
\bottomrule
\end{tabular}
\end{table*}

\subsection{Diversity and Rewritten Synthetic Mixtures Do Not Add Benefit}
\label{sec:diversity_results}

\begin{figure}[t]
\centering
\includegraphics[width=0.95\linewidth]{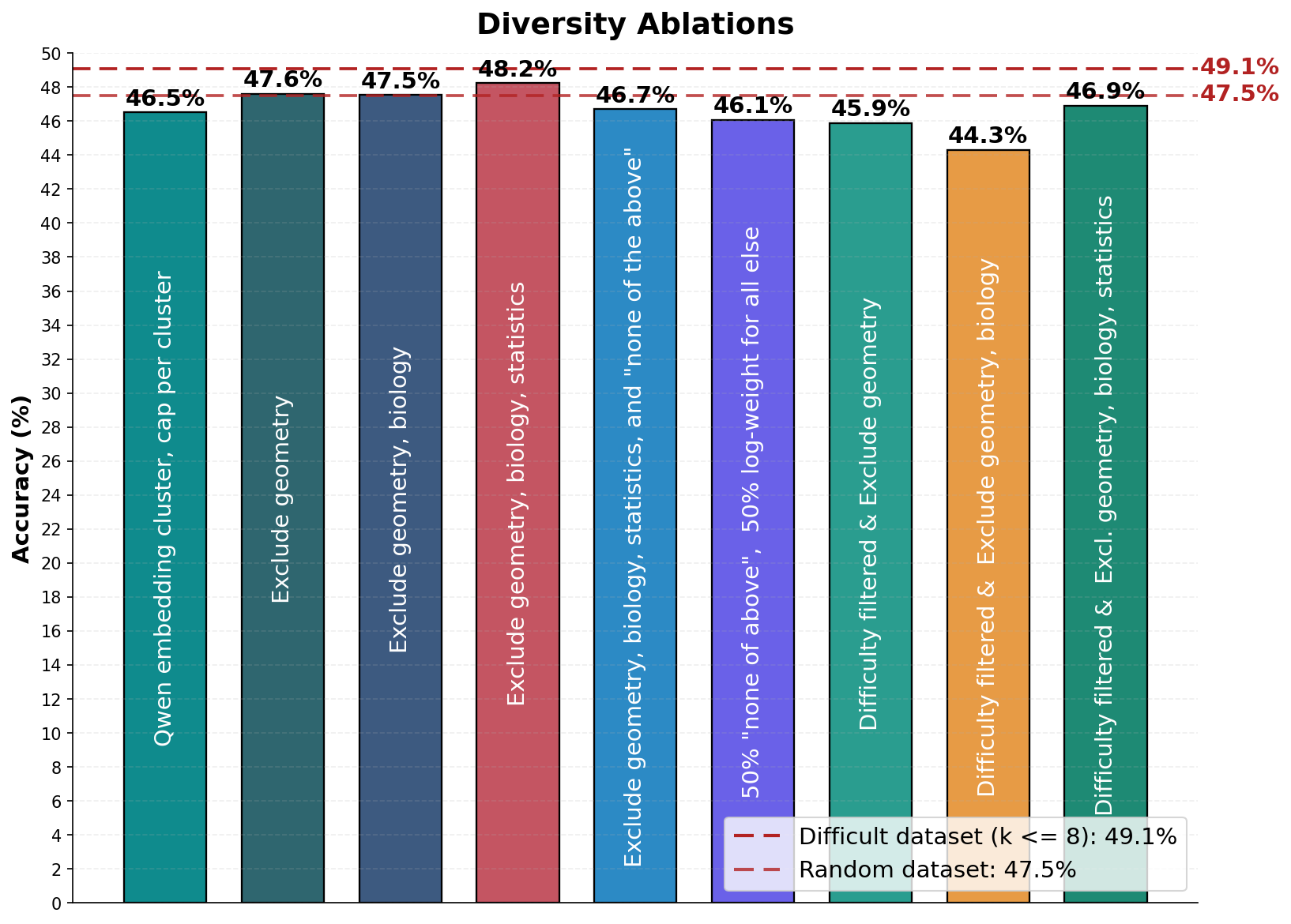}
\caption{Diversity-oriented ablations on Walton. Representative clustering and category-balancing heuristics do not improve over the difficulty-filtered baseline under the fixed DCVLR recipe.}
\label{fig:diversity_ablations}
\end{figure}

The diversity and synthetic-data results are negative but scoped. We evaluate embedding-cluster balancing, category-level balancing, topic exclusions, and combinations with difficulty filtering. Some diversity variants outperform simple random-style baselines, indicating that the implementations are not vacuous. However, none outperform the aligned difficulty-filtered baseline, and combining diversity constraints with difficulty filtering does not provide additive benefit in our tested configurations (Figure~\ref{fig:diversity_ablations}). On Open-R1, the same qualitative pattern holds (Table~\ref{tab:openr1_ablation}): difficulty filtering reaches 46.4, while embedding clustering and topic exclusion each reach 45.3, and difficulty plus topic exclusion reaches 44.4.

\begin{table}[t]
\centering
\caption{Diversity ablations on the Open-R1 base dataset. Scores are overall DCVLR weighted accuracy (\%). Difficulty filtering remains the strongest tested strategy, confirming the Walton finding on a second corpus.}
\label{tab:openr1_ablation}
\setlength{\tabcolsep}{4pt}
\begin{tabular}{lc}
\toprule
Curation strategy & Acc. (\%) \\
\midrule
Difficulty-filtered ($k \le 8$) & \textbf{46.4} \\
Embedding clustering, cap per cluster & 45.3 \\
Topic exclusion: geometry, biology, statistics & 45.3 \\
Difficulty + topic exclusion & 44.4 \\
\bottomrule
\end{tabular}
\end{table}

\begin{figure}[t]
  \centering
  \includegraphics[width=0.95\linewidth]{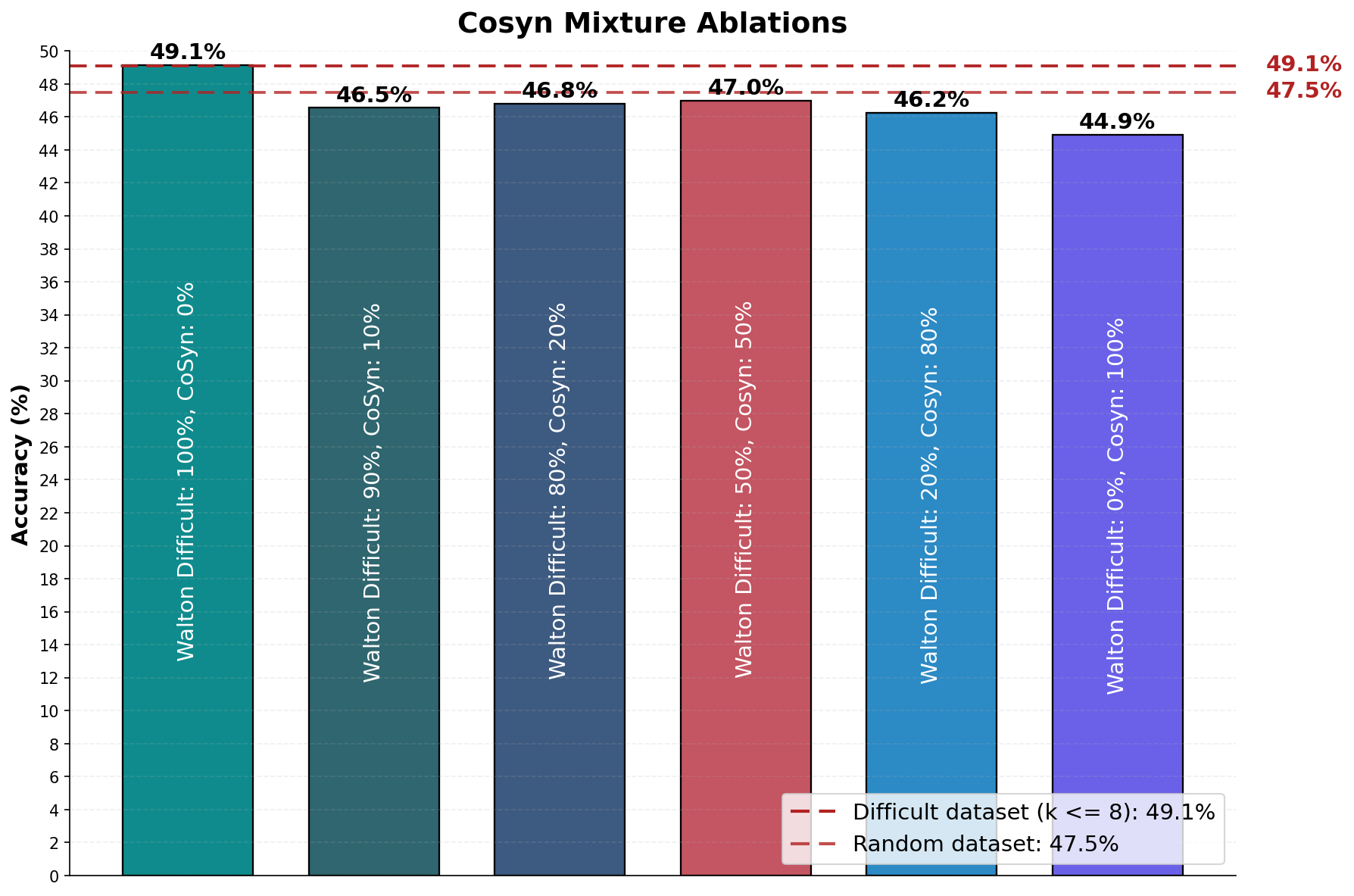}
  \caption{Impact of Walton-to-CoSyn mixture ratios. Increasing the rewritten CoSyn fraction underperforms the Walton-only difficulty-filtered baseline, consistent with distribution or supervision-style mismatch.}
  \label{fig:hard_cosyn_mix}
\end{figure}

Similarly, the CoSyn result should not be read as a claim that synthetic data is generally harmful. Our experiment evaluates a specific mixture: CoSyn examples sampled from the source pool and rewritten into longer reasoning traces while preserving the original final answer. Increasing the fraction of these rewritten examples underperforms the Walton difficulty-filtered baseline (Figure~\ref{fig:hard_cosyn_mix}), which is consistent with distributional or supervision-style mismatch. The result is therefore a caution against assuming that synthetic mixtures help automatically, not a rejection of synthetic data as a general strategy.

\section{Practical Guidance and Scope of Claims}
\label{sec:discussion}

\paragraph{A practical recipe for fixed-protocol curation.}
Our results suggest a simple workflow for data-constrained multimodal reasoning fine-tuning when the model and training recipe are fixed. First, choose the most aligned available source corpus, because source alignment determines whether small curated subsets can provide efficient gains. Second, score examples using a model-relative difficulty signal and select a broad challenging-but-learnable region rather than only the hardest failures. Third, treat dataset size as a stability knob once mean performance saturates; adding more data may reduce variance without improving average accuracy. Finally, evaluate diversity and synthetic augmentation through matched-scale ablations instead of assuming that they compose additively with difficulty filtering.

\paragraph{Scope of claims.}
The central claim of this paper is not that difficulty is universally dominant in all multimodal data curation. The claim is narrower: under the fixed DCVLR-style fine-tuning recipe, difficulty is the strongest signal among the interventions we tested once an aligned source dataset is chosen. Likewise, our diversity and synthetic-data results are negative findings about the representative heuristics and rewritten CoSyn mixtures we evaluated, not universal claims that diversity or synthetic data cannot help. Our dataset-size result says that additional Walton data did not improve mean accuracy under this fixed recipe, not that scale is unimportant in pretraining, mid-training, or unconstrained post-training.

\paragraph{Limitations.}

This study is intentionally scoped to a fixed-protocol fine-tuning regime. We do not vary the optimizer, learning-rate schedule, number of training steps, or main base model in the DCVLR experiments, and therefore cannot determine whether larger or more diverse datasets would be more useful under separately tuned recipes. Our primary analysis starts from Walton, an aligned source corpus; conclusions may differ when only less-aligned source data are available. The difficulty metric is model-relative, and cross-model experiments show partial but not universal transfer. Finally, several diversity, CoSyn, and cross-model checks are single-run or low-repetition experiments due to evaluation cost. These limitations mean that our results should be read as guidance for efficient DCVLR-style curation, not as universal prescriptions for all stages of multimodal training.

\paragraph{Acknowledgments}\mbox{}\\
This work was supported by the National Science Foundation NRT-AI 2244574.

This work was enabled in part by funding from the National Science Foundation award: 2149133.

This work used cloud GPU resources at NCSA Delta cluster through allocation number CIS240027 from the Advanced Cyberinfrastructure Coordination Ecosystem: Services \& Support (ACCESS) program, which is supported by National Science Foundation grants \# 2138259, \# 2138286, \# 2138307, \# 2137603, and \# 2138296.

This work used Jetstream2 at Indiana University through allocation CIS251375 from the Advanced Cyberinfrastructure Coordination Ecosystem: Services \& Support (ACCESS) program, which is supported by National Science Foundation grants \# 2138259, \# 2138286, \# 2138307, \# 2137603, and \# 2138296.

The technical support and advanced computing resources from University of Hawaii Information Technology Services – Research Cyberinfrastructure, funded in part by the National Science Foundation CC* awards \# 2201428 and \# 2232862 are gratefully acknowledged.

This material is based upon work supported by the National Science Foundation CISE Graduate Fellowships under Grant \# 2313998. Any opinions, findings, and conclusions or recommendations expressed in this material are those of the author(s) and do not necessarily reflect the views of the National Science Foundation.

\bibliographystyle{icml2026}
\bibliography{dcvlr_workshop}

@misc{dcvlr2025resultblog,
  author = {Elachqar, O. and Feuer, B. and Tripathi, R. and Zhang, Y. and Hulkund, N. and Nguyen, T. and Shabtay, N. and Udandarao, V. and Wang, X. and Webb, S. and Koukoumidis, E. and Schmidt, L. and Xie, S. and Yeung-Levy, S. and Liang, P. and Beery, S. and Gkioxari, G.},
  title = {DCVLR competition results: Data curation for vision-language reasoning},
  year = {2025},
  howpublished = {Oumi Blog},
  url = {https://blog.oumi.ai/p/dcvlr-competition-results-data-curation}
}

@misc{dcvlr2025website,
  author = {{DCVLR Organizers}},
  title = {Data Curation for Vision--Language Reasoning (DCVLR)},
  year = {2025},
  howpublished = {Project website},
  url = {https://dcvlr-neurips.github.io/}
}

@misc{oumi2025trainingfrontiervlms,
  author = {{Oumi}},
  title = {Training frontier reasoning VLMs},
  year = {2025},
  howpublished = {Blog post},
  url = {https://blog.oumi.ai/p/training-frontier-reasoning-vlms}
}

@misc{oumi2025waltondataset,
  author = {{Oumi}},
  title = {Walton Multimodal Cold Start (R1 format)},
  year = {2025},
  howpublished = {Hugging Face dataset},
  url = {https://huggingface.co/datasets/oumi-ai/walton-multimodal-cold-start-r1-format}
}

@misc{oumi2025mmopenr1,
  author = {{Oumi}},
  title = {Multimodal Open R1 8192 (filtered, mid-IC)},
  year = {2025},
  howpublished = {Hugging Face dataset},
  url = {https://huggingface.co/datasets/oumi-ai/multimodal-open-r1-8192-filtered-mid-ic}
}

@misc{oumi2025mmmathinstruct,
  author = {{Oumi}},
  title = {MM-MathInstruct to R1 format (filtered)},
  year = {2025},
  howpublished = {Hugging Face dataset},
  url = {https://huggingface.co/datasets/oumi-ai/MM-MathInstruct-to-r1-format-filtered}
}

@misc{penfever2025datapreproc,
  author = {Penfever},
  title = {data-preproc},
  year = {2025},
  howpublished = {GitHub repository},
  url = {https://github.com/penfever/data-preproc}
}

@misc{shabtay2025livexivmultimodallive,
  title = {LiveXiv -- A Multi-Modal Live Benchmark Based on ArXiv Papers Content},
  author = {Shabtay, N. and Polo, F. M. and Doveh, S. and Lin, W. and Mirza, M. J. and Chosen, L. and Yurochkin, M. and Sun, Y. and Arbelle, A. and Karlinsky, L. and Giryes, R.},
  year = {2025},
  eprint = {2410.10783},
  archivePrefix = {arXiv},
  primaryClass = {cs.CV},
  url = {https://arxiv.org/abs/2410.10783}
}

@misc{bai2025qwen25vltechnicalreport,
  title = {Qwen2.5-VL Technical Report},
  author = {Bai, S. and Chen, K. and Liu, X. and Wang, J. and Ge, W. and Song, S. and Dang, K. and Wang, P. and Wang, S. and Tang, J. and Zhong, H. and Zhu, Y. and Yang, M. and Li, Z. and Wan, J. and Wang, P. and Ding, W. and Fu, Z. and Xu, Y. and Ye, J. and Zhang, X. and Xie, T. and Cheng, Z. and Zhang, H. and Yang, Z. and Xu, H. and Lin, J.},
  year = {2025},
  eprint = {2502.13923},
  archivePrefix = {arXiv},
  primaryClass = {cs.CV},
  url = {https://arxiv.org/abs/2502.13923}
}

@misc{zhang2025vmcbench,
  title = {Automated Generation of Challenging Multiple-Choice Questions for Vision Language Model Evaluation},
  author = {Zhang, Y. and Su, Y. and Liu, Y. and Wang, X. and Burgess, J. and Sui, E. and Wang, C. and Aklilu, J. and Lozano, A. and Wei, A. and others},
  year = {2025},
  eprint = {2501.03225},
  archivePrefix = {arXiv},
  primaryClass = {cs.CL}
}

@misc{marsili2025omni3dbench,
  title = {Visual Agentic AI for Spatial Reasoning with a Dynamic API},
  author = {Marsili, D. and Agrawal, R. and Yue, Y. and Gkioxari, G.},
  year = {2025},
  eprint = {2502.06787},
  archivePrefix = {arXiv},
  primaryClass = {cs.CV}
}

@misc{feng2025physics,
  title = {Physics: Benchmarking Foundation Models on University-Level Physics Problem Solving},
  author = {Feng, K. and Zhao, Y. and Liu, Y. and Yang, T. and Zhao, C. and Sous, J. and Cohan, A.},
  year = {2025},
  eprint = {2503.21821},
  archivePrefix = {arXiv},
  primaryClass = {cs.CL}
}

@misc{yang2025scaling,
  author = {Yang, Y. and Patel, A. and Deitke, M. and Gupta, T. and Weihs, L. and Head, A. and Yatskar, M. and Callison-Burch, C. and Krishna, R. and Kembhavi, A.},
  title = {Scaling text-rich image understanding via code-guided synthetic multimodal data generation},
  year = {2025},
  eprint = {2502.14846},
  archivePrefix = {arXiv},
  primaryClass = {cs.CV}
}

@misc{gekhman2024finetuning,
  title = {Does Fine-Tuning LLMs on New Knowledge Encourage Hallucinations?},
  author = {Gekhman, Z. and Yona, G. and Aharoni, R. and Eyal, M. and Feder, A. and Reichart, R. and Herzig, J.},
  year = {2024},
  eprint = {2405.05904},
  archivePrefix = {arXiv},
  primaryClass = {cs.CL},
  url = {https://arxiv.org/abs/2405.05904}
}

\clearpage
\appendix

\section{Comparison of DCVLR Submission Strategies}
\label{app:submissions}

\begin{table*}[t]
\centering
\scriptsize
\setlength{\tabcolsep}{4pt}
\begin{tabularx}{\textwidth}{l l l l l l c X}
\toprule
\textbf{Team} &
\textbf{Base} &
\textbf{Curation} &
\textbf{Difficulty} &
\textbf{Diversity} &
\textbf{Distill.} &
\textbf{Size} &
\textbf{Keywords} \\
\midrule
Team 1 & Multi-source & Filtering & Model failure rate & Implicit (source) & gpt-4o / VLM CoT & 1K & Difficult filtering, failure-based \\
Team 2 & Multi-source & Filtering & Category-balanced & Category sampling & Multi-LLM judges & $\sim$10K & Reject sampling, refinement \\
Team 3 & Images & Synthetic & Reward score & Implicit (images) & Qwen2.5-VL & $\sim$10K & Synthetic instr., reward \\
Team 4 & Multi-source & Filtering & Fail-score (72B) & Dedup + img-dep. & Qwen3-235B CoT & 10K & Fail ranking, image-grounded \\
Team 5 & Open + synth. & Hybrid & Task tiering & Domain balance & gpt-5 / Gemini & 10K & Multi-domain, curriculum \\
Team 6 & Multi-source & Filtering & Error mining & Category balance & Gemini-2.5 Pro & $\sim$10K & Weak-model failures \\
Team 7 & Existing & Filtering & CoT complexity & Embedding clusters & Existing CoTs & 100 / 1K & Complexity + diversity \\
\bottomrule
\end{tabularx}
\caption{Dataset curation strategies of top DCVLR submissions. Most favor difficulty-aware filtering with expert reasoning distillation over large-scale synthetic generation.}
\label{tab:dcvlr_curation_comparison}
\end{table*}

Curation strategy details in Table~\ref{tab:dcvlr_curation_comparison} are derived from publicly available competition write-ups and system description cards released by participating teams on the DCVLR leaderboard. Teams are anonymized in order of final ranking. No private communications were used.

\section{Alignment and Pipeline Details}
\label{app:alignment}

\begin{figure}[h]
\centering
\includegraphics[width=0.95\linewidth]{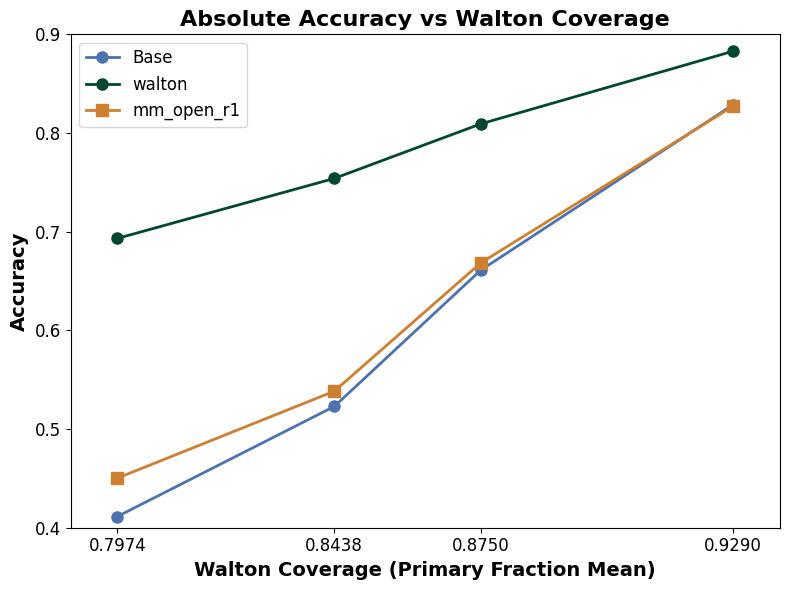}
\caption{Absolute accuracy on LiveXivTQA vs. Walton coverage, defined as the fraction of Walton examples among kNN neighbors.}
\label{fig:walton_coverage}
\end{figure}

For each LiveXivTQA example, we compute $k$-nearest neighbors ($k=32$) against the union of Walton and MM-Open-R1 embeddings extracted from the base Qwen2.5-VL-7B-Instruct model. We measure the fraction of neighbors from Walton and bin questions by this fraction. Figure~\ref{fig:walton_coverage} shows that the base model is more accurate on LiveXivTQA questions that are more Walton-like, supporting the interpretation that Walton is aligned with both the benchmark and the base model.

\section{Implementation and Reproducibility Details}
\label{app:impl}

\paragraph{Stochastic decoding.}
Difficulty scores are computed with 16 rollouts at temperature 0.7 and top-$p$ 0.9. We used 16 rollouts for analysis resolution, but preliminary checks with 8 rollouts produced qualitatively similar coarse filtering behavior. The full 16-rollout scoring pass requires approximately $16\times$ inference over the Walton corpus.

  \paragraph{Answer matching.}
  Correctness is determined by a three-stage process applied to each response. \emph{Stage 1} compares the stripped prediction to the ground truth by exact, case-insensitive, and whitespace-normalized matching (no punctuation or numeric-format normalization). \emph{Stage 2} attempts structured extraction in priority order: LaTeX \texttt{\textbackslash boxed\{\}} (nested-brace aware, last box), math/tag expressions, SymPy symbolic equivalence (\texttt{simplify(pred-gt)==0}), and natural-language answer spans; an extracted span is accepted only if it matches the ground-truth length, unless both sides parse to non-integer floats. Multiple-choice and boolean extractors are enabled only when the ground-truth answer is short (\textless 15 characters). \emph{Stage 3} routes unresolved cases to an LLM judge (Qwen3-4B) that extracts the model's answer and rules on semantic or mathematical equivalence, with the choice options supplied for multiple-choice benchmarks.

\paragraph{CoSyn sampling and rewriting.}
We draw 1{,}000 CoSyn-400K examples by uniform random sampling (seed 42) from the first 10{,}240 examples of the \texttt{validation} split, using the \texttt{data-preproc} tool~\citep{penfever2025datapreproc}, and resize images to at most $768\times768$ with aspect ratio preserved; no deduplication, token-length or image-count filtering is applied at this stage. Each example is then rewritten with GPT-4o (\texttt{gpt-4o-2024-08-06}, temperature 0), conditioned on the image, the question, and the ground-truth answer, into a step-by-step reasoning trace whose final answer is re-emitted in \texttt{\textbackslash boxed\{\}}. The rewriting instruction was:

\begingroup\footnotesize
\begin{verbatim}
Rewrite a clear, step-by-step reasoning trace.
Ensure the final answer is EXACTLY the given
ground-truth answer.
Put the final answer within \boxed{...}.
\end{verbatim}
\endgroup

The rewritten traces are mixed with difficulty-filtered Walton at controlled ratios (e.g., 10\%), using each rewritten trace as the training target while preserving the original question, image and final answer.

\paragraph{Preprocessing.}
Final datasets are constructed with the organizer-provided \texttt{data-preproc} tool~\citep{penfever2025datapreproc}, including deduplication, length-based filtering, and format normalization.

\paragraph{Repetition counts.}
Random 1k Walton uses five repetitions. Difficulty thresholds use three repetitions per setting. Dataset-size ablations use four repetitions at 250 examples, five at 1k, four at 5k, three at 7.5k, and one at 10k. Cross-model transfer, CoSyn mixtures, and most diversity variants are reported as single-run or low-repetition checks unless explicitly marked.

\paragraph{Compute and release.}
Difficulty scoring for the full Walton corpus required approximately 200 GPU-hours on A100-class GPUs. Fine-tuning runs followed the official DCVLR protocol. We plan to release code, difficulty scores, random seeds, and curated data identifiers where licensing permits.

\section{Clustering and Category-Balancing Details}
\label{app:diversity}

\begin{figure*}[t]
\centering
\includegraphics[width=\linewidth]{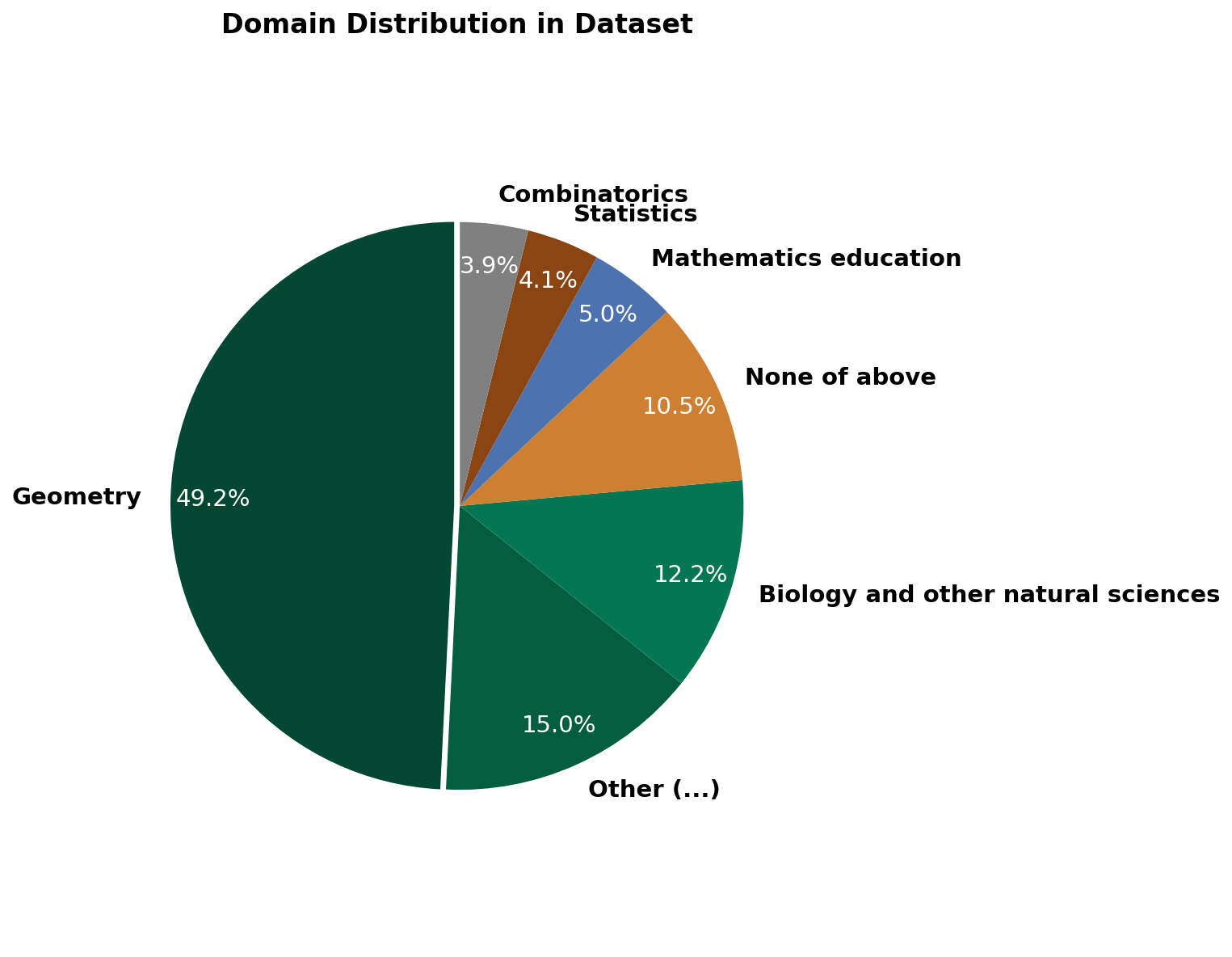}
\caption{Distribution of Mathematics Subject Classification (MSC) categories in the Walton dataset.}
\label{fig:msc_pie}
\end{figure*}

Each Walton example is embedded using the base Qwen2.5-VL-7B-Instruct model. Embeddings are clustered with $k$-means, then sampled using square-root-proportional allocation with per-cluster caps and Dirichlet-based randomization. We swept cluster granularity, caps, and selection ratios. For category balancing, each Walton example is assigned a coarse Mathematics Subject Classification category (Figure~\ref{fig:msc_pie} shows the resulting distribution); we sweep category exclusions, per-domain caps, and log-weighted sampling. While some configurations outperform random selection, none approach the difficulty-filtered baseline under the fixed recipe.

\end{document}